\documentclass[sigconf]{acmart}
\copyrightyear{2026}
\acmYear{2026}
\setcopyright{cc}
\setcctype{by}
\acmConference[WWW '26] {Proceedings of the ACM Web Conference 2026}{April 13--17, 2026}{Dubai, United Arab Emirates.}
\acmBooktitle{Proceedings of the ACM Web Conference 2026 (WWW '26), April 13--17, 2026, Dubai, United Arab Emirates}
\acmISBN{979-8-4007-2307-0/2026/04}
\acmDOI{10.1145/3774904.3792631}

\usepackage{algorithm}
\usepackage{algpseudocode}
\usepackage{multirow}
\usepackage{booktabs}
\usepackage{graphicx}
\usepackage{enumitem}

\setlist[itemize]{leftmargin=1em, itemsep=0pt, parsep=0pt, topsep=2pt}

\begin{document}

\title{MetaKube: An Experience-Aware LLM Framework for Kubernetes Failure Diagnosis}

\author{Wei Sun}
\authornote{Both authors contributed equally to this research.}
\affiliation{%
  \institution{School of Science and Engineering\\The Chinese University of Hong Kong, Shenzhen\\Shenzhen Future Network of Intelligence Institute}
  \city{Shenzhen}
  \country{China}}
\email{weisun2@link.cuhk.edu.cn}

\author{Ting Wang}
\authornotemark[1]
\affiliation{%
  \institution{School of Science and Engineering\\The Chinese University of Hong Kong, Shenzhen\\Shenzhen Future Network of Intelligence Institute}
  \city{Shenzhen}
  \country{China}}
\email{tingwang@link.cuhk.edu.cn}

\author{Xinran Tian}
\affiliation{%
  \institution{Shenzhen Future Network of Intelligence Institute}
  \city{Shenzhen}
  \country{China}}
\email{tianxinran1022@gmail.com}

\author{Wanshun Lan}
\affiliation{%
  \institution{China Mobile Communications Group Guangdong Co., Ltd.}
  \city{Guangzhou}
  \country{China}}
\email{lanwanshun@gd.chinamobile.com}

\author{Xuhan Feng}
\affiliation{%
  \institution{China Mobile Communications Group Guangdong Co., Ltd.}
  \city{Guangzhou}
  \country{China}}
\email{fengxuhan@gd.chinamobile.com}

\author{Haoyue Li}
\affiliation{%
  \institution{China Mobile Communications Group Guangdong Co., Ltd.}
  \city{Guangzhou}
  \country{China}}
\email{13802887950@139.com}

\author{Fangxin Wang}
\authornote{Corresponding author.}
\affiliation{%
  \institution{School of Science and Engineering\\The Chinese University of Hong Kong, Shenzhen\\Shenzhen Future Network of Intelligence Institute}
  \city{Shenzhen}
  \country{China}}
\email{wangfangxin@cuhk.edu.cn}

\renewcommand{\shortauthors}{Wei Sun et al.}


\begin{abstract}
Existing LLM-based Kubernetes diagnostic systems cannot learn from operational experience, operating on static knowledge bases without improving from past resolutions. We present MetaKube, an experience-aware LLM framework through three synergistic innovations: (1) an Episodic Pattern Memory Network (EPMN) that abstracts diagnostic patterns from historical resolutions and provides confidence-calibrated retrieval for both rapid pattern matching and guided causal exploration, (2) a meta-cognitive controller that dynamically routes between intuitive and analytical pathways based on problem familiarity, optimizing the trade-off between speed and depth, and (3) KubeLLM, a locally-deployable 8B model enhanced through domain-specific post-training on our 7,000-sample Kubernetes Fault Resolution Dataset. Evaluation on 1,873 real-world scenarios demonstrates MetaKube transforms Qwen3-8B from 50.9 to 90.5 points, approaching GPT-4.1 performance while ensuring complete data privacy. EPMN contributes 15.3\% improvement through experiential learning, with continuous learning experiments showing progressive gains as the system accumulates operational knowledge. The source code and related resources are available at https://github.com/MetaKube-LLM-for-Kubernetes-Diagnosis/MetaKube.
\end{abstract}


\maketitle

\section{Introduction}

Kubernetes \cite{burns2016borg} has become the critical infrastructure orchestrating containerized applications across global cloud deployments. However, its operational complexity in production environments poses significant challenges for fault diagnosis \cite{barletta2024criticality}. Modern clusters contain thousands of containers with intricate dependencies across pods, services, and custom resources, making it extremely difficult for operators to trace failure propagation and identify root causes from scattered monitoring data \cite{kannaiah2024kubernetes}.

Large Language Models (LLMs) present transformative potential for Kubernetes fault diagnosis through their reasoning capabilities and knowledge integration \cite{ghorab2025detection}. Recent work \cite{guu2020retrieval} \cite{zhang2024llm} explores retrieval-augmented generation (RAG) to enhance LLMs with domain-specific knowledge, enabling conversational troubleshooting interfaces that interpret error messages and recommend remediation strategies. These initial efforts demonstrate promising results in reducing the expertise threshold for cluster management \cite{tippins2024domain}.

However, deploying solutions based on LLMs in production Kubernetes environments faces three fundamental challenges \cite{boateng2025survey}. \textbf{First}, existing RAG systems cannot learn from operational experience—they operate on static knowledge repositories without feedback mechanisms \cite{li2024eaco}, treating each diagnostic session in isolation and ignoring valuable insights from historical resolutions. \textbf{Second}, the Kubernetes ecosystem suffers from acute scarcity of high-quality diagnostic data \cite{hrusto2025monitoring}, with troubleshooting knowledge fragmented across documentation, GitHub issues, and proprietary runbooks, severely constraining both model training and retrieval effectiveness. \textbf{Third}, enterprise deployments require strict data privacy, prohibiting external API usage for sensitive cluster information \cite{gupta2024api}. This creates a difficult trade-off: large open-source LLMs (70B+ parameters) impose prohibitive computational overhead, while smaller LLMs (<10B parameters) lack the specialized reasoning capabilities needed for accurate diagnostics \cite{savage2025open,almubark2025exploring}.

To address these fundamental challenges, we present MetaKube, a cognitive architecture that synergistically integrates episodic memory networks, specialized language models, and causal knowledge graphs within a unified diagnostic framework. MetaKube implements dual reasoning pathways: an intuitive pathway leveraging our novel Episodic Pattern Memory Network (EPMN) for rapid pattern recognition, and an analytical pathway utilizing KubeGraph for systematic causal exploration. A meta-cognitive controller employing confidence-calibrated assessment dynamically orchestrates pathway selection based on problem complexity and memory similarity scores. \textbf{Our contributions are threefold:}

\begin{itemize}
\item\textbf{First}, we propose the MetaKube architecture to address the critical deficiency of learning from operational experience. MetaKube integrates three synergistic components: (1) an Episodic Pattern Memory Network (EPMN) that continuously abstracts and refines diagnostic patterns from resolution outcomes, (2) a specialized language model (KubeLLM) that adapts through post-training, and (3) a dynamic knowledge graph (KubeGraph) that expands its causal structure through discovered relationships, collectively enabling the system to improve diagnostic accuracy through operational feedback.

\item\textbf{Second}, to overcome the extreme scarcity of high-quality diagnostic data in the Kubernetes ecosystem, we construct comprehensive diagnostic resources by aggregating and structuring knowledge from diverse online sources including documentation, forums, and issue trackers. This effort produces both a Kubernetes knowledge graph encompassing causal relationships and operational constraints, and a curated fault resolution dataset covering common failure patterns.

\item\textbf{Third}, considering the stringent data privacy requirements in production Kubernetes environments, we adapt the locally deployable Qwen3-8B~\cite{yang2025qwen3} model as MetaKube's backbone, employing advanced post-training techniques—specifically domain-specific supervised fine-tuning on our large-scale curated Kubernetes Fault Resolution Dataset—to significantly elevate its diagnostic reasoning and resolution capabilities to levels comparable to commercial API services. This approach ensures complete on-premise deployment, keeping sensitive cluster data securely within organizational boundaries while delivering state-of-the-art diagnostic performance.
\end{itemize}

Experiments on KubeFault, our dataset of 1,873 Kubernetes fault scenarios, validate MetaKube's effectiveness: it improves Qwen3-8B by 40.6 points, approaching GPT-4.1 performance while ensuring data privacy. 

\section{Related Work}

\subsection{Kubernetes Fault Diagnosis}

Traditional Kubernetes fault diagnosis approaches range from rule-based systems to machine learning techniques \cite{kaul2020ai,boluda2025enhancing}. Statistical methods like CloudRanger~\cite{8411065} correlate metrics for root cause analysis, while graph-based approaches including MicroRCA~\cite{9110353} trace anomaly propagation patterns. Commercial platforms (Datadog, Dynatrace) and open-source tools (Prometheus, Jaeger) provide comprehensive monitoring but lack semantic understanding of failure causality. Recent AIOps systems~\cite{8802836} combine multiple signals yet remain fundamentally reactive, requiring extensive manual feature engineering and struggling with novel failure modes. These approaches collectively fail to capture the adaptive learning capabilities demonstrated by expert operators who continuously refine their diagnostic techniques based on prior experiences \cite{fritzsche2024lightweight}. MetaKube transcends these limitations through episodic memory that continuously learns diagnostic patterns from operational experience, eliminating predefined rules while adapting to emerging failure modes.

\subsection{LLMs for System Operations}

Recent work explores LLMs for system administration, with OpsGPT~\footnote{Broadridge Financial Solutions, Inc., ``Broadridge Launches GenAI-Powered OpsGPT to Transform and Optimize Trading in a T+1 Environment,'' press release, January 11, 2024, \url{https://www.broadridge.com/press-release/2024/broadridge-launches-genai-powered-opsgpt}.} automating incident mitigation and X-Lifecycle~\cite{goel2024x} enabling conversational cloud management. Studies by ~\cite{maeno2024panda} reveal both promise and limitations in operational contexts. Advanced prompting techniques like chain-of-thought~\cite{wei2022chain} and ReAct~\cite{yao2023react} improve diagnostic reasoning but cannot accumulate experience across sessions. While these general-purpose models generate semantically plausible solutions, they cannot guarantee operational feasibility in production environments, potentially suggesting configurations incompatible with specific Kubernetes versions or cluster constraints \cite{carrion2022kubernetes,furnadzhiev2025efficient}. Additionally, concerns about data privacy arise when sensitive system logs and error messages are processed by external LLM services, posing significant risks for enterprise deployments \cite{malik2024securing,rathod2025privacy}. These privacy concerns are particularly acute in enterprise environments where regulatory compliance and data sovereignty requirements often preclude sending sensitive operational data to external LLM services \cite{miller2025leveraging,hong2025survey}. KubeLLM addresses these gaps as the first foundation model specifically engineered for Kubernetes diagnostics, combining specialized training on curated troubleshooting corpora with continuous adaptation, while maintaining on-premise deployment options for sensitive environments.

\subsection{Retrieval-Augmented Generation Systems}

RAG architectures enhance LLMs with external knowledge, from REALM's~\cite{guu2020retrieval} end-to-end retrieval training to RETRO's~\cite{borgeaud2022improving} trillion-token scaling. Self-RAG~\cite{asai2023self} introduces adaptive retrieval decisions, while K8sGPT and InfAgent~\cite{zhang2024llm} apply RAG to Kubernetes operations. However, these systems face three critical limitations: they retrieve from static knowledge bases without incorporating operational feedback, lack mechanisms to abstract patterns from successful diagnostic episodes, and cannot distinguish effective strategies from failed attempts \cite{gao2025survey,wu2025position}. Additionally, current approaches rely on similarity-based retrieval that ignores temporal dynamics and confidence assessment in diagnostic contexts \cite{zhang2024failure}. MetaKube addresses these limitations through memory-augmented dual-pathway processing that transforms static retrieval into adaptive pattern recognition with continuous learning capabilities.

\begin{figure*}[h]
   \centering
   \includegraphics[width=0.95\textwidth]{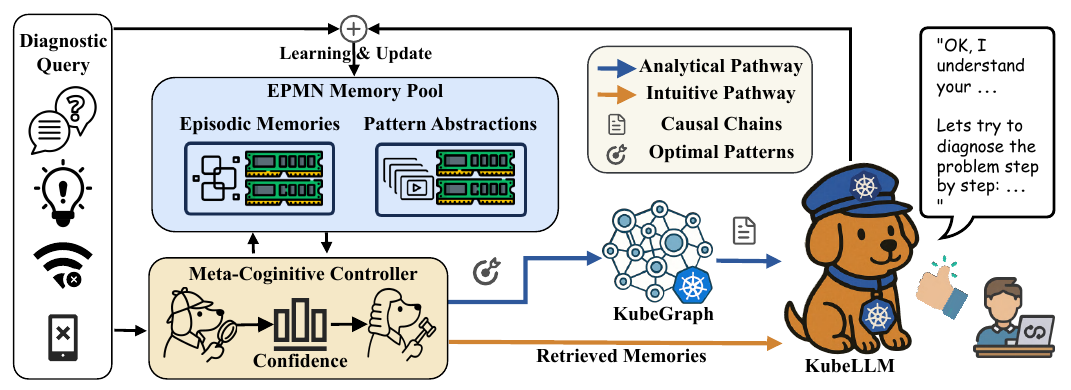}
   \caption{The MetaKube dual-pathway diagnostic architecture. The system integrates memory-augmented pattern recognition (EPMN) with causal reasoning (KubeGraph) through two processing pathways: an intuitive pathway for rapid pattern-based diagnosis and an analytical pathway for complex causal analysis. A meta-cognitive controller dynamically routes queries based on confidence assessment, optimizing the trade-off between diagnostic speed and depth. }\label{fig:main}
\end{figure*}

\section{MetaKube: An Experience-Aware LLM Framework for Kubernetes
Failure Diagnosis}\label{3}

This section presents MetaKube, a cognitive-inspired diagnostic framework that integrates memory-augmented pattern recognition \cite{santoro2016meta} with causal reasoning. As illustrated in Figure~\ref{fig:main}, the architecture comprises dual processing pathways orchestrated by a meta-cognitive controller for adaptive Kubernetes diagnostics.

\subsection{Overview and Design Philosophy}

MetaKube implements cognitive dual-process theory \cite{evans2013dual} for Kubernetes diagnostics, emulating expert operators' ability to transition between pattern recognition and analytical reasoning. Our architecture integrates two memory-augmented pathways: an intuitive pathway employing EPMN for rapid pattern retrieval and solution synthesis via KubeLLM, and an analytical pathway combining EPMN with KubeGraph's causal reasoning capabilities. The innovation lies in memory augmentation across both pathways, where episodic patterns simultaneously enable rapid responses and provide heuristics for efficient causal exploration.

A meta-cognitive controller dynamically routes queries based on confidence assessment derived from memory similarity, consistency, and recency metrics. When confidence exceeds threshold $\tau$, the system activates the computationally efficient intuitive pathway; otherwise, it triggers the analytical pathway for thorough causal investigation. This adaptive routing optimizes computational resources proportionate to problem complexity.

The architecture enables continuous knowledge-experience co-evolution through bidirectional learning. Diagnostic outcomes refine memory patterns via adaptive weighting, while causal reasoning validates empirical knowledge by revealing underlying mechanisms. This self-improvement cycle transcends static retrieval approaches by continuously integrating operational experience into both recognition and reasoning capabilities.

\subsection{Dual-Pathway Processing Flow}

Both the intuitive pathway $\mathcal{P}_{int}$ and analytical pathway $\mathcal{P}_{ana}$ accept diagnostic query vectors $\mathcal{Q} \in \mathbb{R}^d$ encoding symptom descriptions and real-time cluster state, yet they diverge fundamentally in their computational strategies and depth of insight.

\subsubsection{Intuitive Pathway}

The intuitive pathway $\mathcal{P}_{int}$ embodies experience driven diagnosis, enabling immediate response through sophisticated pattern matching. When presented with a diagnostic query $\mathcal{Q} = \{s_1, s_2, ..., s_n\}$, the pathway initiates a streamlined processing sequence:

\begin{equation}
\mathcal{P}_{int}: \mathcal{Q} \xrightarrow{\text{EPMN}} \mathcal{M}^* \xrightarrow{\text{KubeLLM}_{\theta_{int}}} \mathcal{S}_{int}
\end{equation}

EPMN performs intelligent memory retrieval, selecting optimal patterns $\mathcal{M}^* \subseteq \mathcal{M}_{pool}$ from the comprehensive memory pool $\mathcal{M}_{pool} = \mathcal{E} \cup \mathcal{P}$ by optimizing:

\begin{equation}
\mathcal{M}^* = \arg\max_{ |\mathcal{M}| \leq K} 
\sum_{m_i \in \mathcal{M}} 
\left[\lambda \cdot \text{sim}(m_i, \mathcal{Q}) + (1-\lambda) \cdot \text{rec}(m_i)\right]
\end{equation}

where similarity $\text{sim}(m_i, \mathcal{Q})$ employs cosine similarity in the embedding space and recency $\text{rec}(m_i) = \exp(-\Delta t_i/\tau_r)$ introduces temporal awareness. The balance parameter $\lambda \in [0,1]$ allows dynamic adjustment between semantic relevance and temporal recency.

Subsequently, KubeLLM operates in template mode with parameters $\theta_{int}$, transforming retrieved patterns into coherent diagnostic solutions. This configuration leverages historical patterns as contextual anchors, enabling rapid inference grounded in proven solutions. The pathway's activation depends on aggregate confidence:

\begin{equation}
C(\mathcal{M}^*) = \max_{m \in \mathcal{M}^*} C(m, \mathcal{Q})
\end{equation}

This mechanism ensures the intuitive pathway handles recurring issues like resource exhaustion and scheduling failures with millisecond-scale responses while maintaining diagnostic accuracy through experiential validation.

\subsubsection{Analytical Pathway}

The analytical pathway $\mathcal{P}_{ana}$ extends beyond pattern recognition to perform memory-guided causal reasoning for complex or unprecedented failures. Building upon the same memory retrieval, this pathway incorporates systematic knowledge graph exploration:

\begin{equation}
\mathcal{P}_{ana}: \mathcal{Q} \xrightarrow{\text{EPMN}} \mathcal{M}^* \xrightarrow{\text{KubeGraph}} \mathcal{G}^* \xrightarrow{\text{KubeLLM}_{\theta_{ana}}} \mathcal{S}_{ana}
\end{equation}

KubeGraph receives retrieved memories $\mathcal{M}^*$ as navigational guides for traversing the knowledge graph $G_K = (V, E, \mathcal{A})$. Rather than exhaustive exploration, the system employs memory-biased search:

\begin{equation}
\text{priority}(p) = \alpha_1 \cdot \text{prior}(\mathcal{M}^*, p) + 
\alpha_2 \cdot \text{score}_{path}(p) + \alpha_3 \cdot \text{novelty}(p)
\end{equation}
The prior function $\text{prior}(\mathcal{M}^*, p) = \max_{m \in \mathcal{M}^*} \text{overlap}(p, \text{Path}(m))$ quantifies historical support. Through multi-hop expansion guided by these priorities, the search extracts causal chains $\mathcal{G}^* = \{c_1, ..., c_n\}$ where each chain represents a validated causal relationship.

KubeLLM then operates in analytical mode ($\theta_{ana}$), synthesizing comprehensive diagnostics that integrate memory patterns with discovered causal evidence. The resulting solution $\mathcal{S}_{ana}$ provides complete reasoning chains with explicit causal attribution, enabling operators to understand not just what failed but why.

\subsubsection{Unified Processing Framework}

The dual-pathway architecture maintains three fundamental invariants: completeness ($\mathcal{S}_{int} \subseteq \mathcal{S}_{ana}$), ensuring analytical solutions encompass intuitive diagnoses; monotonicity ($C(\mathcal{S}_{ana}) \geq C(\mathcal{S}_{int})$), reflecting that deeper analysis strengthens confidence; and convergence, where intuitive solutions progressively approach analytical quality for recurring problems.

The formal diagnostic function maps inputs through dynamically selected pathways:
\begin{equation}
\mathcal{F}: (\mathcal{Q}, \Theta) \xrightarrow{\Psi} 
\begin{cases}
\mathcal{P}_{int} \to (\mathcal{S}_{int}, C_{int}) & \text{if } C(\mathcal{M}^*) > \tau \\
\mathcal{P}_{ana} \to (\mathcal{S}_{ana}, C_{ana}) & \text{otherwise}
\end{cases}
\end{equation}

This design ensures simple problems receive rapid solutions while complex issues trigger comprehensive analysis, optimizing both resource utilization and diagnostic quality.

\subsection{Meta-Cognitive Control and Optimization}\label{meta}

The meta-cognitive controller $\Psi$ transcends simple pathway selection to implement continuous learning and optimization. This meta-layer embodies cognitive monitoring functions that enable the system to gauge problem familiarity and adapt reasoning strategies accordingly.

The controller implements a sophisticated mapping:
\begin{equation}
\Psi: (\mathcal{Q}, \mathcal{M}^*, \Theta) \to (p, \tau, \omega)
\end{equation}

where query $\mathcal{Q}$ and retrieved patterns $\mathcal{M}^*$ inform pathway selection $p \in \{\mathcal{P}_{int}, \mathcal{P}_{ana}\}$, while simultaneously adapting confidence threshold $\tau$ and optimization parameters $\omega$.

Confidence assessment integrates multiple dimensions of diagnostic relevance:
\begin{equation}
C(\mathcal{M}^*) = \max_{m \in \mathcal{M}^*} \prod_{j} f_j(m, \mathcal{Q})^{\nu_j}
\end{equation}

Factors include similarity $f_{sim} = \exp(-d(\mathcal{Q}, m)/\sigma_{sim})$, temporal relevance $f_{temp} = \exp(-\Delta t/T_{temp})$, historical success $f_{succ}$, and context compatibility $f_{ctx}$. Weights $\nu_j$ undergo continuous adaptation through gradient-based optimization.

The routing strategy implements confidence-driven branching:
\begin{equation}
\mathcal{P}(\mathcal{Q}) = 
\begin{cases}
\mathcal{P}_{int} & \text{if } C(\mathcal{M}^*) > \tau \\
\mathcal{P}_{ana} & \text{if } C(\mathcal{M}^*) \leq \tau
\end{cases}
\end{equation}

The threshold $\tau$ itself adapts through meta-learning to balance accuracy and efficiency:
\begin{equation}
\tau_{t+1} = \tau_t - \eta_{meta} \nabla_{\tau} L(\tau_t, \mathcal{H})
\end{equation}
where the loss function captures the fundamental trade-off:
\begin{equation}
L(\tau, \mathcal{H}) = \xi \cdot \text{Error}(\tau, \mathcal{H}) + (1-\xi) \cdot \text{Latency}(\tau, \mathcal{H})
\end{equation}

This creates a self-optimizing system that learns its own limitations and continuously refines its understanding of when to trust experience versus engage deeper analysis. Successful intuitive diagnoses gradually lower the threshold to expand rapid response coverage, while errors raise it to ensure quality.

\section{Core Components and Synergistic Design}

The realization of MetaKube's dual-process architecture depends fundamentally on three core components that work in concert to enable cognitive diagnosis. Building upon the architectural framework established in Section \ref{3}, we now examine how EPMN provides experiential learning, KubeGraph enables structural reasoning, and KubeLLM synthesizes natural language understanding. Each component contributes unique capabilities while their integration creates emergent diagnostic intelligence that transcends individual contributions.

\subsection{EPMN: Episodic Pattern Memory Network}\label{sec:epmn}

The Episodic Pattern Memory Network serves as MetaKube's experiential foundation, transforming diagnostic history into actionable intelligence. As established in our dual-pathway design, EPMN generates memory patterns $\mathcal{M}^*$ that guide both rapid intuitive response and systematic analytical exploration.

\subsubsection{Dual-Granularity Memory Architecture}

EPMN implements a dual-layer memory structure $\mathcal{M}_{EPMN} = \{\mathcal{E}, \mathcal{P}, f_{extract}, f_{recall}\}$ that maintains both specific episodic memories and generalized pattern abstractions. Each episode $e_i \in \mathcal{E}$ encodes a complete diagnostic trajectory as $e_i = \langle s_i, c_i, a_i, o_i, t_i, \omega_i \rangle$, capturing symptoms, context, actions, outcomes, temporal marking, and adaptive memory value.

The pattern abstraction layer $\mathcal{P}$ emerges through clustering similar experiences:
\begin{equation}
p_j = f_{abstract}(\{e_i | \text{sim}(e_i, e_k) > \theta_{sim}\})
\end{equation}

Each pattern $p_j = \{\mu_j^s, \Sigma_j^s, \phi_j, \rho_j\}$ captures statistical properties of symptom clusters ($\mu_j^s$, $\Sigma_j^s$), canonical resolution strategy ($\phi_j$), and reliability metrics ($\rho_j$). The similarity threshold $\theta_{sim}$ controls pattern granularity, enabling adaptive balance between generalization and specificity.

\subsubsection{Confidence-Aware Retrieval}

When processing query $\mathcal{Q}$, EPMN performs parallel search across both memory layers:
\begin{equation}
\mathcal{M}^* = \text{TopK}(\psi \cdot \mathcal{M}_{\mathcal{P}} \cup (1-\psi) \cdot \mathcal{M}_{\mathcal{E}}, K)
\end{equation}

The mixing parameter $\psi$ adapts based on query characteristics:
\begin{equation}
\psi = \sigma(W_{\psi} \cdot [f_{nov}(\mathcal{Q}), f_{comp}(\mathcal{Q})])
\end{equation}
where novelty $f_{nov}(\mathcal{Q}) = \min_{m \in \mathcal{M}} d(\mathcal{Q}, m)$ measures distance to nearest historical experience, and complexity $f_{comp}(\mathcal{Q}) = H(s_{\mathcal{Q}})$ quantifies symptom entropy. Novel or complex queries favor specific episodes; familiar patterns leverage abstracted knowledge.

Multi-factor confidence computation provides meta-cognitive routing signals:
\begin{equation}
C(m, \mathcal{Q}) = \prod_{j} f_j(m, \mathcal{Q})^{\zeta_j}
\end{equation}
The weights $\zeta_j$ undergo specialized optimization for confidence calibration, ensuring estimates align with actual diagnostic difficulty.

\subsubsection{Component Integration}

EPMN enhances other components through targeted knowledge sharing. For KubeGraph, it provides exploration hints:
\begin{equation}
\text{Hints} = \{v \in V_G | \exists m \in \text{TopK}_{hint}(\mathcal{Q}): v \in \text{Path}(m)\}
\end{equation}

These hints focus graph traversal on historically relevant paths, reducing computational overhead while maintaining diagnostic completeness. For KubeLLM, retrieved patterns enrich reasoning context through structured injection.
Meta-cognitive integration relies on comprehensive confidence signals:
\begin{equation}
\text{Signal}_{meta} = \{C_{max}, C_{avg}, C_{std}, \text{Coverage}\}
\end{equation}

When routing decisions contradict confidence estimates, the system learns through calibration loss:
\begin{equation}
\mathcal{L}_{calibration} = \text{BCE}(C_{predicted}, \mathbb{I}[\text{fast\_sufficient}])
\end{equation}

Computational efficiency scales logarithmically through hierarchical indexing: $\mathcal{O}(\log|\mathcal{P}| + K_{rel}\log|\mathcal{E}|)$. Under stationary failure distributions, pattern abstraction converges to stable diagnostic archetypes while maintaining adaptability through continuous episodic learning.

The detailed procedural implementation of these mechanisms is presented in Algorithm \ref{alg:epmn} of Appendix~\ref{1}.

\subsection{KubeLLM: Context-Aware Diagnostic Synthesizer}

Our backbone model, KubeLLM, comprises two essential components that collectively form a comprehensive framework for Kubernetes problem diagnosis. First, we detail the pipeline for constructing a specialized Kubernetes troubleshooting dataset, capturing the unique characteristics of real-world diagnostic scenarios. Second, we implement Supervised Fine-Tuning (SFT) as a cold-start mechanism to inject domain-specific Kubernetes knowledge and cultivate structured reasoning capabilities within the model.

\subsubsection{Kubernetes Fault Resolution Dataset}
\label{sec:data_preprocess}

We constructed the Kubernetes Fault Resolution Dataset (KFRD), the first dataset specifically collected and built for Kubernetes fault resolution. Our multi-stage pipeline systematically processes real-world Kubernetes issues to create a comprehensive training corpus.

\begin{figure}[h]
   \centering
   \includegraphics[width=0.45\textwidth]{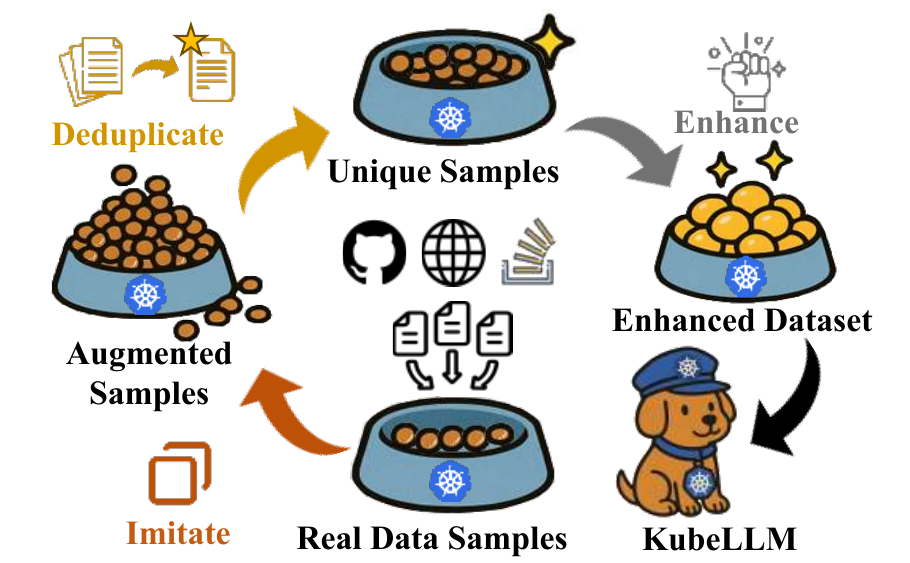}
   \caption{{Construction of the Kubernetes Fault Resolution Dataset. Generate additional synthetic examples after collecting authentic Kubernetes troubleshooting cases, then performed deduplication, and enhanced the question-answer pairs with chain-of-thought reasoning using LLM.}}\label{fig:gou}
\end{figure}

The dataset construction follows four key stages: (1) \textbf{Problem-Solution Collection} from technical forums including Stack Overflow and GitHub issues; (2) \textbf{Problem-Attempt-Solution Reformulation} to capture the distinctive pattern where users document failed attempts before seeking help; (3) \textbf{Data Augmentation} using LLMs to expand the dataset while maintaining structural integrity; and (4) \textbf{Chain-of-Thought Enhancement} to enrich solutions with explicit reasoning paths. The complete pipeline is illustrated in Figure \ref{fig:gou}.

Our final dataset comprises 7,000 high-quality samples, strategically partitioned into 5,000 for SFT, and 2,000 for evaluation. The structured format ensures that even synthetic examples maintain logical flow and contextual richness characteristic of genuine troubleshooting interactions. (Detailed construction methodology and formulations are provided in Appendix~\ref{app:dataset_construction}.)

\subsubsection{Supervised Fine-Tuning with Low-Rank Adaptation}
\label{sec:sft}

While foundation models possess baseline Kubernetes knowledge, their performance on specialized diagnostic tasks remains suboptimal. We employ supervised fine-tuning (SFT) with Low-Rank Adaptation (LoRA) to efficiently inject domain-specific knowledge while preserving foundational capabilities.

Our approach adapts the pre-trained Qwen3-8B model using structured problem-solution pairs from KFRD. The LoRA mechanism decomposes weight updates into low-rank matrices:

\begin{equation}
W = W_0 + \Delta W = W_0 + A \cdot B^T
\end{equation}

where $W_0$ represents frozen pre-trained weights, and $A \in \mathbb{R}^{d \times r}$, $B \in \mathbb{R}^{k \times r}$ with rank $r \ll \min(d,k)$. This significantly reduces trainable parameters from $\mathcal{O}(d \cdot k)$ to $\mathcal{O}(r(d + k))$, enabling efficient adaptation while minimizing catastrophic forgetting.

The SFT phase serves dual purposes: enhancing Kubernetes diagnostic capabilities and providing a robust component as part of MetaKube. (Complete training objectives and implementation details are described in Appendix~\ref{app:sft_details}.)

\begin{table*}[t]
\centering
\caption{Overall performance comparison on KubeFault dataset (100-point scale).}
\label{tab:overall_results}
\resizebox{\textwidth}{!}{
\begin{tabular}{l|ccccc|ccccc}
\hline
\multirow{2}{*}{\textbf{Method}} & \multicolumn{5}{c|}{\textbf{GPT-5 Automated Assessment}} & \multicolumn{5}{c}{\textbf{Human Expert Evaluation}} \\
\cline{2-11}
& \textbf{Eff.} & \textbf{Equ.} & \textbf{Com.} & \textbf{S/A} & \textbf{Avg.} & \textbf{Eff.} & \textbf{Equ.} & \textbf{Com.} & \textbf{S/A} & \textbf{Avg.} \\
\hline
GPT-4.1 (Zero-shot) & 72.1±4.2 & 74.3±3.8 & 69.8±4.5 & 78.9±3.1 & 73.8±2.9 & 56.4±9.1 & 59.2±9.8 & 52.7±9.4 & 64.8±8.2 & 58.3±7.3 \\
GPT-4.1-mini (Zero-shot) & 61.5±5.1 & 63.8±4.6 & 59.2±5.4 & 69.7±4.2 & 63.6±3.7 & 45.1±10.3 & 47.6±10.9 & 41.8±10.1 & 55.4±9.5 & 47.5±8.4 \\
Qwen3-8B (Zero-shot) & 48.7±6.8 & 51.2±6.3 & 46.1±7.1 & 57.4±5.7 & 50.9±5.2 & 31.5±12.1 & 35.8±11.6 & 28.9±12.4 & 42.3±10.8 & 34.6±9.7 \\
GPT-4.1 (GraphRAG) & 89.3±2.9 & \textbf{92.6±2.4} & \textbf{91.4±2.7} & \textbf{94.1±2.3} & \textbf{91.9±1.8} & 73.8±7.3 & \textbf{77.8±7.9} & \textbf{71.2±8.2} & 79.4±7.1 & \textbf{75.6±6.1} \\
GPT-4.1-mini (GraphRAG) & 79.8±4.1 & 81.3±3.7 & 78.4±4.5 & 85.2±3.2 & 81.2±2.9 & 59.3±9.1 & 61.7±8.8 & 57.4±9.4 & 68.9±8.3 & 61.8±7.2 \\
Qwen3-8B (GraphRAG) & 66.7±5.4 & 69.1±5.1 & 64.8±5.8 & 73.3±4.6 & 68.5±3.8 & 44.2±10.6 & 47.8±9.9 & 42.1±10.4 & 54.6±9.8 & 47.2±8.6 \\
\hline
MetaKube (Ours) & \textbf{91.2±2.4} & 90.8±2.6 & 87.3±2.9 & 92.5±2.2 & 90.5±2.0 & \textbf{75.6±8.1} & 74.2±8.5 & 69.8±7.8 & \textbf{81.2±7.2} & 75.2±6.4 \\
\hline
\end{tabular}
}
\begin{flushleft}
\footnotesize
\textit{Eff.}: Effectiveness; \textit{Equ.}: Equivalence; \textit{Com.}: Completeness; \textit{S/A}: Safety/Accuracy; \textit{Avg.}: Average score across all dimensions.
\end{flushleft}
\end{table*}

\subsection{KubeGraph: Reliable Knowledge Base}
\label{sec:kg_construction}

To enhance the reliability of the MetaKube system, we constructed a specialized knowledge graph for Kubernetes operations. Unlike traditional knowledge graph construction methods that rely on complex manual or semi-automatic NLP pipelines, we leverage emerging approaches exemplified by GraphRAG\cite{edge2024local}, utilizing LLMs to efficiently generate knowledge graphs from unstructured text. Our construction process encompasses three core phases:

\subsubsection{Corpus Curation}
We systematically collected comprehensive K8s operational knowledge from authoritative sources to build a robust text corpus $\mathcal{T}$. These sources are strategically categorized into three main types: (1) \textbf{Official documentation and community resources}, including Kubernetes.io and reference blogs from authoritative operational companies (e.g., NewRelic, Alibaba Cloud), ensuring knowledge accuracy; (2) \textbf{Technical blogs and articles} from platforms such as StackOverflow and Medium, incorporating real-world cases and solutions; and (3) \textbf{Professional books} including ``Kubernetes in Action,'' and ``Kubernetes Practice Guide,'' establishing systematic knowledge frameworks.

This multi-source approach ensures our knowledge foundation integrates both theoretical principles and practical operational experiences, creating a reliable and comprehensive knowledge base for the K8s domain.

\subsubsection{LLM-based Pre-processing and Categorization}
Direct application of raw text to GraphRAG would introduce significant semantic noise and factual inconsistencies. To address this challenge, we designed and implemented an LLM-based automated preprocessing workflow:

\begin{equation}
    f_{\text{preprocess}}: \mathcal{T} \xrightarrow[]{\text{LLM}} \mathcal{D}_{kg} = \{(d_j, \kappa_j)\}_{j=1}^{N}
\end{equation}
where $\mathcal{T}$ represents the original corpus, and $\mathcal{D}_{kg}$ represents the processed dataset containing $N$ document-category pairs. Each document undergoes rigorous content cleaning to remove noise elements (navigation bars, advertisements, user comments) and precise topic categorization through carefully engineered prompts to the GPT-4.1 API.

The topic classification task assigns each cleaned document to one of seven predefined categories:

\begin{equation}
    \kappa_j = \text{LLM}_{\text{classify}}(\hat{d}_j) \in \mathcal{K} = \{\kappa_1, \kappa_2, ..., \kappa_7\}
\end{equation}
where $\mathcal{K}$ represents the category set: ``K8s Explanations and Introductions,'' ``Resource Errors,'' ``Network Errors,'' ``Scheduling Errors,'' ``Image Errors,'' ``Configuration Errors,'' and ``System Errors.'' These categories comprehensively cover the most common fault domains in K8s operations.

To validate the reliability of this automated process, we conducted a thorough evaluation with experienced K8s operations engineers. The validation results confirmed that our process achieved high standards in both classification accuracy and content fidelity, establishing a foundation for high-quality knowledge graph generation.

All processed data was structured in JSON format, including additional metadata such as source, timestamp, and confidence scores for further analysis and graph construction. This systematic approach ensures the resulting knowledge graph contains only verified, reliable information that provides crucial support for the entire problem-solving workflow. Detailed KubeGraph specifications and performance metrics are provided in Appendix~\ref{appendix:kubegraph}.

\section{Experiments}

\subsection{Experimental Setup}
\textbf{Dataset.} We evaluate MetaKube using \textbf{KubeFault}, a collection of 1,873 Kubernetes fault scenarios extracted from KubeGraph using GPT-5. These scenarios span six error categories (Resource, Network, Scheduling, Image, Configuration, System Errors), with symptom descriptions, environmental context, logs, ground truth root causes, and resolution suggestions all automatically annotated. To ensure authority and accuracy, all generated content was reviewed and corrected by experienced operations engineers from telecommunications companies.

\textbf{Baselines.} We compare against four GraphRAG-based approaches: GPT-4.1 (zero-shot), GPT-4.1-mini (zero-shot), Qwen3-8B (zero-shot),GPT-4.1 (GraphRAG), GPT-4.1-mini (GraphRAG), Qwen3-8B (GraphRAG). The GraphRAG variants leverage graph-structured knowledge retrieval but lack MetaKube's specialized components like episodic memory and dynamic knowledge graphs.

\textbf{Metrics.} Our evaluation uses metrics across four dimensions: Effectiveness: measures solution completeness and root cause resolution capability; Equivalence: assesses alignment with reference approaches and methodology consistency; Completeness: evaluates coverage of necessary steps, commands, and edge cases; Safety/Accuracy: examines correctness and adherence to Kubernetes best practices. To ensure fair and objective evaluation, we employ two parallel scoring systems: (1) GPT-5 automated assessment and (2) blind evaluation by three experienced operations engineers from different telecommunications companies with scores averaged. This dual-metric approach provides both scalable automated evaluation and expert human validation.

\textbf{Implementation details} are provided in Appendix~\ref{appendix:implementation}.

\subsection{Comparative Evaluation}

We present a comprehensive comparison of MetaKube against baseline methods across the KubeFault dataset using a 100-point evaluation scale. Table~\ref{tab:overall_results} summarizes the performance across four evaluation dimensions using both GPT-5 automated assessment and human expert evaluation. Our MetaKube method builds upon the Qwen3-8B foundation model, optimized specifically for Kubernetes environments through our KubeLLM component.

MetaKube demonstrates exceptional performance, achieving a remarkable 40.6-point improvement over the base Qwen3-8B model (90.5 vs. 50.9 in GPT-5 assessment) and reaching within 1.4 points of GPT-4.1 GraphRAG while operating on significantly smaller computational resources. This transformation of a modest open-source model into a highly capable Kubernetes fault diagnosis system validates our multi-agent architecture and domain-specific optimization approach. In human expert evaluation, MetaKube excels particularly in critical operational dimensions, achieving the highest effectiveness score and safety/accuracy rating, demonstrating that our approach delivers practical value that operations engineers can trust in production environments.

The performance gains stem from our specialized KubeGraph construction and multi-agent coordination, which capture the complex interdependencies and failure propagation patterns inherent in Kubernetes environments. While graph-enhanced approaches consistently outperform zero-shot methods across all models, MetaKube's unique combination of domain-specific knowledge representation, coordinated reasoning agents, and targeted optimization delivers superior results while maintaining complete data privacy for enterprise deployments. The consistent performance across both evaluation methodologies underscores the robustness of our framework and its alignment with real-world operational needs.

\begin{figure}[t]
	\centering
	\includegraphics[width=0.45\textwidth]{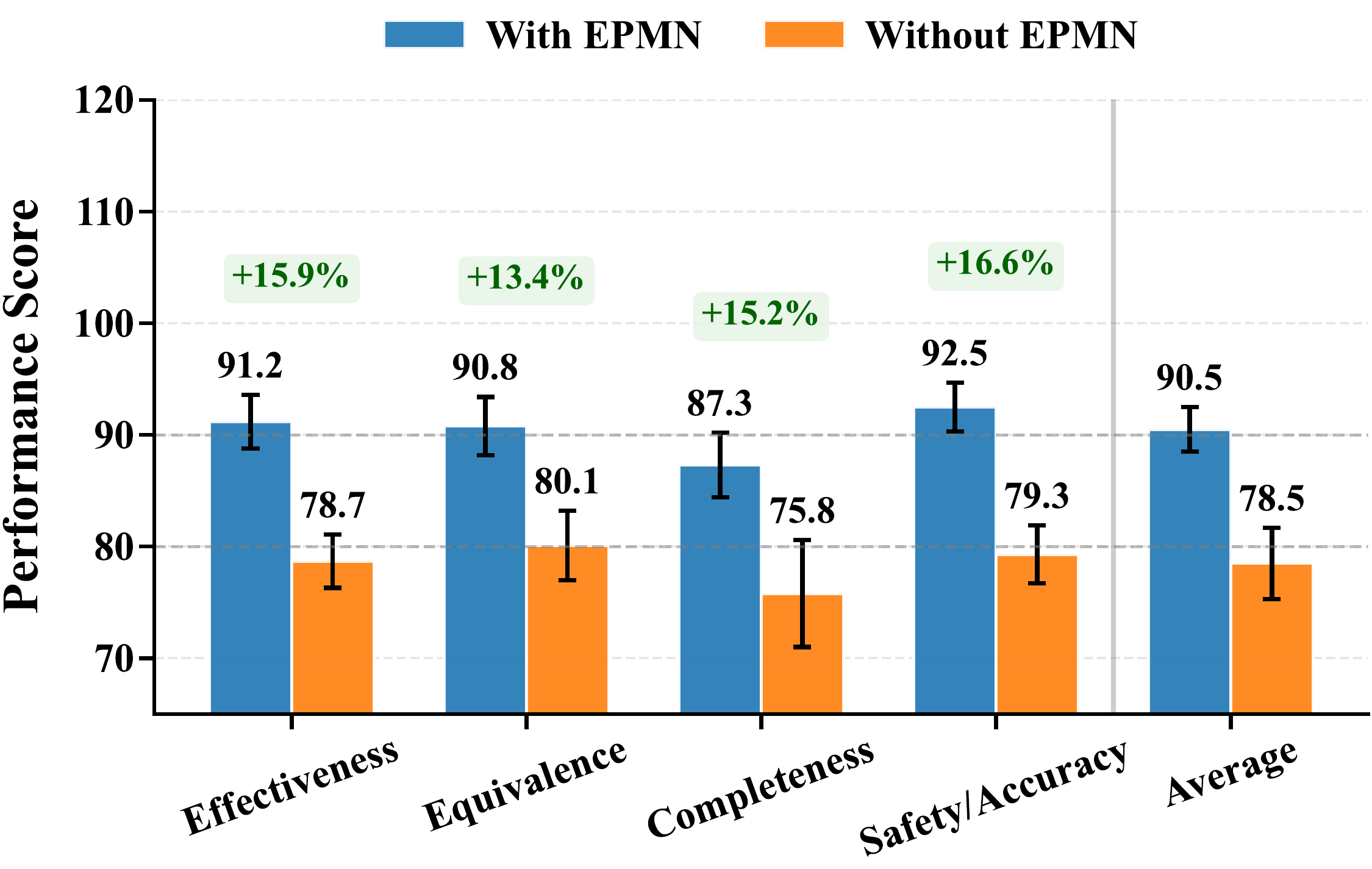}
	\caption{EPMN ablation study results on KubeFault.}
	\label{fig:epmn_ablation}
\end{figure}

\subsection{Ablation Studies}

\textbf{EPMN Ablation.}
To validate the contribution of our proposed EPMN module, we conducted an ablation study comparing system performance with and without the EPMN component. The evaluation was performed across four key metrics: Effectiveness, Equivalence, Completeness, and Safety/Accuracy, with each experiment repeated five times to ensure statistical reliability.

Figure~\ref{fig:epmn_ablation} demonstrates that the EPMN module provides substantial and consistent improvements across all evaluation dimensions, with performance gains ranging from 13.4\% to 16.6\% and an overall improvement of 15.3\%. The consistent enhancement across all metrics reveals a key advantage of our approach: many underlying problems share common patterns and root causes, allowing the EPMN's memory network to flexibly allocate and reuse learned representations across different scenarios. This pattern consistency enables efficient knowledge transfer and reduces redundant learning, while the dynamic memory mechanism can adaptively retrieve and combine relevant patterns based on input characteristics. The particularly notable improvement in Safety/Accuracy (16.6\%) demonstrates the EPMN's superior ability to capture and address consistent error patterns through memory-guided processing, making it highly suitable for practical kubernetes deployment scenarios where reliability is paramount.

\begin{figure*}[t]
	\centering
	\includegraphics[width=0.9\textwidth]{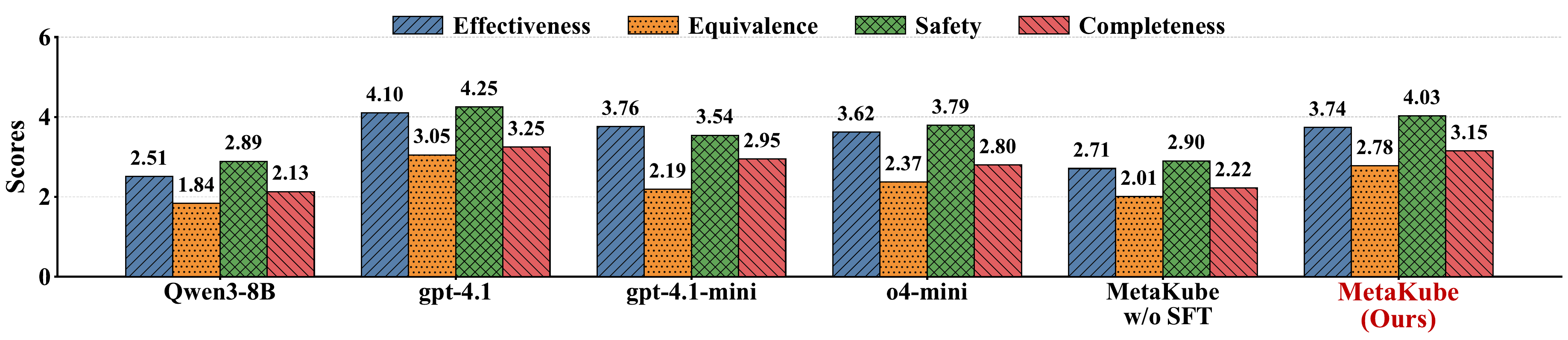}
	\caption{Performance scores of different models on the KFRD test set (5-point scale).}
	\vspace{-3mm}
	\label{fig:model_comparison}
\end{figure*}

\textbf{KubeLLM Ablation.}
As MetaKube's backbone, KubeLLM handles intent interpretation, knowledge understanding, and language generation, making enhancement of the base model (Qwen3-8B) essential. In this set of experiments, we evaluate the improvement in accuracy on the KFRD validation set before and after SFT training, and demonstrate that an 8B-parameter model can approximate the correctness achieved by mainstream API-based solutions in our Kubernetes fault resolution scenario (see Figure~\ref{fig:model_comparison}). Given that most Kubernetes fault issues do not have a unique solution, we assess model responses along four dimensions: \textbf{Effectiveness} (resolution of root causes), \textbf{Equivalence} (alignment with reference solutions), \textbf{Completeness} (coverage of all necessary steps), and \textbf{Safety} (adherence to K8s best practices). We perform multiple evaluations of the instructions using the GPT-5 API as judge, and report the average score as the final metric. The results indicate that, under these objective evaluation criteria, our backbone model exhibits a 45.5\% performance improvement after the SFT training.

\textbf{KubeGraph Ablation.}
To further validate the effectiveness of our KubeGraph component, we conducted comprehensive ablation studies on both in-domain and out-of-domain datasets. The evaluation was performed on KubeFault (in-domain) and a real-world Kubernetes fault problem dataset provided by a telecommunications company (out-of-domain), with performance assessed through expert scoring across the same four metrics.

\begin{table}[h]
\centering
\caption{KubeGraph Ablation Results.}
\resizebox{\columnwidth}{!}{%
\begin{tabular}{lccccc}
\toprule
\textbf{Dataset} & \textbf{Eff.} & \textbf{Equ.} & \textbf{Com.} & \textbf{S/A} & \textbf{Avg.} \\
\midrule
\textbf{KubeFault } & & & & & \\
w/ KubeGraph & 75.6±8.1 & 74.2±8.5 & 69.8±7.8 & 81.2±7.2 & 75.2±6.4 \\
w/o KubeGraph & 31.5±12.1 & 35.8±11.6 & 28.9±12.4 & 42.3±10.8 & 34.6±9.7 \\
\midrule
\textbf{Telecom Dataset } & & & & & \\
w/ KubeGraph & 58.4±9.2 & 55.7±9.6 & 52.3±8.9 & 64.1±8.4 & 57.6±7.8 \\
w/o KubeGraph & 19.8±13.5 & 24.2±12.8 & 17.1±13.7 & 28.5±11.9 & 22.4±11.2 \\
\bottomrule
\end{tabular}%
}
\end{table}

The results reveal the critical importance of KubeGraph for both generalization and domain-specific performance. On in-domain KubeFault data, KubeGraph provides substantial improvements of 117.3\% overall, with particularly strong gains in Completeness (141.5\%) and Effectiveness (140.0\%). More importantly, KubeGraph demonstrates superior generalization capabilities on out-of-domain telecom data with 157.1\% overall improvement, indicating that KubeGraph's structural representation learning captures fundamental problem-solving patterns that transcend specific domains, leveraging structural similarities and topological patterns that remain consistent across different technical environments.

\section{Conclusion}

We presented MetaKube, a memory-augmented architecture that transforms Kubernetes diagnosis through continuous experiential learning. Our key innovation lies in the synergistic integration of episodic pattern memory, dual-pathway processing, and meta-cognitive control, enabling systems to accumulate and leverage operational knowledge across diagnostic sessions. MetaKube achieves 90.5 points on real-world scenarios, approaching GPT-4 performance (91.9) while ensuring complete data privacy through on-premise deployment. The 15.3\% improvement from memory augmentation and progressive gains through continuous learning demonstrate the critical importance of experience-based learning in operational AI. Our comprehensive release of KFRD, KubeGraph, and KubeLLM addresses fundamental data scarcity challenges and provides the community with essential building blocks for advancing experience-aware diagnosis in distributed systems. 

\section*{Acknowledgments}

The work was supported in part by the National Key Research and Development Program of China (Grant No. 2024YFB2907000), the Basic Research Project No. HZQB-KCZYZ-2021067 of Hetao Shenzhen-HK S\&T Cooperation Zone, the Guangdong S\&T Programme (Grant No. 2024B0101030002), the National Natural Science Foundation of China(Grant No. 62293482 and Grant No. 62471423), the Shenzhen Science and Technology Program (Grant No. JCYJ2024 1202124021028 and Grant No. JCYJ20230807114204010), the Guangdong Talents Program (Grant No. 2024TQ08X346), the Shenzhen Outstanding Talents Training Fund 202002, the Young Elite Scientists Sponsorship Program of CAST (Grant No. 2022QNRC001), the Guangdong Provincial Key Laboratory of Future Networks of Intelligence (Grant No. 2022B1212010001) and the Shenzhen Key Laboratory of Big Data and Artificial Intelligence (Grant No. SYSPG20241211173853027).

\bibliographystyle{ACM-Reference-Format}
\bibliography{sample-base}

\appendix

\section{Experimental Implementation Details}\label{appendix:implementation}

\subsection{Dataset Construction}

\textbf{KubeFault Dataset Generation}
The KubeFault dataset comprises 1,873 Kubernetes fault scenarios systematically extracted from production environments, distributed across six error categories: Resource Errors (412, 22.0\%), Network Errors (387, 20.7\%), Scheduling Errors (298, 15.9\%), Image Errors (276, 14.7\%), Configuration Errors (315, 16.8\%), and System Errors (185, 9.9\%). Each fault scenario includes symptom descriptions in natural language (averaging 150 tokens), environmental context capturing cluster configuration and resource states, log traces from kubectl outputs truncated to 2048 tokens, verified root causes with causal chain explanations, and step-by-step resolution suggestions with executable kubectl commands.

\textbf{Data Validation Process}
Three senior operations engineers from telecommunications companies independently reviewed all generated scenarios, achieving inter-rater agreement of Cohen's $\kappa = 0.87$ \cite{finch2018educational}. Discrepancies were resolved through consensus discussion, resulting in 312 command syntax corrections, 178 root cause refinements, and 96 resolution step reorderings to ensure technical accuracy and operational validity.

\subsection{Baseline Implementations}

\textbf{Zero-Shot Baselines}
Zero-shot baselines (GPT-4.1, GPT-4.1-mini, Qwen3-8B) utilize direct prompting without retrieval augmentation, employing a structured prompt template requesting root cause analysis, resolution steps, and prevention recommendations based on provided symptoms, context, and logs.

\textbf{GraphRAG Baselines}
GraphRAG implementations leverage Microsoft GraphRAG framework v0.3.2 \cite{edge2024local}, constructing knowledge graphs from Kubernetes documentation and 10K historical incidents. The configuration employs text-embedding-3-large for embeddings, 512-token chunks with 50-token overlap, top-K=10 subgraph retrieval, and Leiden community detection with resolution=1.0.

\subsection{MetaKube Configuration}

\textbf{Model Components}
EPMN operates with a memory pool of 5,000 episodes, pattern threshold $\theta_{sim}=0.85$, retrieval size K=10, and temporal decay $\tau_r=30$ days. KubeGraph encompasses 12 node types (Pod, Service, Node, etc.) and 8 edge types (depends\_on, manages, etc.), with maximum hop distance of 3 and priority weights $(\alpha_1, \alpha_2, \alpha_3)=(0.5, 0.3, 0.2)$. KubeLLM builds upon Qwen3-8B-Instruct with LoRA rank 256 and maximum context length of 8192 tokens. The meta-controller initializes with threshold $\tau_0=0.75$, meta-learning rate $\eta_{meta}=0.01$, and balance parameter $\xi=0.6$.

\textbf{Training Procedure}
KubeLLM is obtained through supervised fine-tuning (SFT) of the Qwen3-8B base model using the KFRD dataset. The SFT is performed on 10K samples for 5 epochs with learning rate $5\times10^{-5}$, batch size 32, LoRA rank 256, and cosine scheduling with warmup ratio 0.01. The remaining 2K samples, which combine real Kubernetes solution data verified by experienced operations engineers from telecommunications companies, are used for validation and testing. All experiments were conducted on two servers: one with 8 A100 GPUs and another with 8 A6000 GPUs, both running Ubuntu 20.04. This training procedure transforms the general-purpose Qwen3-8B model into our specialized KubeLLM component, which serves as the core LLM module within the MetaKube multi-agent architecture.

\subsection{Evaluation Methodology}

\textbf{Metric Definitions}
Each dimension is scored on a 0-10 scale with specific weighted criteria. Effectiveness combines root cause identification (0.4), resolution quality (0.3), and prevention recommendations (0.3). Equivalence equally weights alignment with reference solutions and methodological consistency. Completeness evaluates step coverage (0.3), command accuracy (0.3), and edge case handling (0.4). Safety/Accuracy balances technical correctness and adherence to Kubernetes best practices equally.

\textbf{Human Evaluation Protocol}
Three evaluators with 5+ years of Kubernetes operations experience independently score each diagnostic output under blind conditions with model identities hidden. Following initial calibration on 50 examples, evaluators assess outputs with score validity requiring standard deviation $\sigma < 1.5$. This dual assessment combining GPT-5 automated scoring and expert human validation ensures both scalable and authoritative evaluation.

\subsection{Experimental Environment}

Experiments were conducted on two Ubuntu 20.04 servers equipped with 8×NVIDIA A100 80GB and 8×NVIDIA A6000 48GB GPUs respectively, with AMD EPYC 7763 64-Core processors and 512GB DDR4 memory. The software stack comprises Python 3.10.12, PyTorch 2.1.0 with CUDA 12.1, and Transformers 4.36.0. All experiments employ fixed random seed 42 for reproducibility. 

\section{Algorithm Details}\label{1}
\begin{algorithm}
	\caption{EPMN: Episodic Pattern Memory Network}
	\label{alg:epmn}
	\begin{algorithmic}[1]
		\Statex \textbf{Input:} Query $\mathcal{Q}$, memories $\mathcal{E}$ and patterns $\mathcal{P}$
		\Statex \textbf{Output:} Relevant memories $\mathcal{M}^*$ with confidence $C_{max}$
		
		\Procedure{FormPatterns}{$\mathcal{E}$}
		\For{$e_i \in \mathcal{E}$}
		\State $E_{sim} \gets \{e_j \in \mathcal{E} \mid sim(e_i, e_j) > \theta_{sim}\}$
		\If{$|E_{sim}| \geq \theta_{pattern}$}
		\State $p_{new}.stats \gets \{\text{Centroid}(E_{sim}), \text{Variance}(E_{sim})\}$
		\State $p_{new}.meta \gets \{\text{Strategy}(E_{sim}), \text{Reliability}(E_{sim})\}$
		\State $\mathcal{P} \gets \text{Update}(\mathcal{P}, p_{new})$
		\EndIf
		\EndFor
		\State \Return $\mathcal{P}$
		\EndProcedure
		
		\Procedure{Retrieve}{$\mathcal{Q}, \mathcal{E}, \mathcal{P}$}
		\State $\psi \gets \sigma(W_{\psi} \cdot [\text{Novelty}(\mathcal{Q}), \text{Complexity}(\mathcal{Q})])$
		\State Score patterns: $\mathcal{M}_{\mathcal{P}} \gets \{(p, C(p, \mathcal{Q})) \mid p \in \mathcal{P}\}$
		\State Score episodes: $\mathcal{M}_{\mathcal{E}} \gets \{(e, C(e, \mathcal{Q})) \mid e \in \mathcal{E}\}$
		\State $\mathcal{M}^* \gets \text{TopK}(\psi \cdot \mathcal{M}_{\mathcal{P}} \cup (1-\psi) \cdot \mathcal{M}_{\mathcal{E}}, K)$
		\State $C_{max} \gets \max_{m \in \mathcal{M}^*} C(m, \mathcal{Q})$
		\State \Return $\mathcal{M}^*, C_{max}$
		\EndProcedure
		
		\Procedure{Confidence}{$m, \mathcal{Q}$}
		\State $C(m, \mathcal{Q}) \gets \text{Similarity}(m, \mathcal{Q})^{\zeta_1} \cdot \text{Relevance}(m, \mathcal{Q})^{\zeta_2} \cdot \text{Reliability}(m)^{\zeta_3}$
		\State \Return $C(m, \mathcal{Q})$
		\EndProcedure
	\end{algorithmic}
\end{algorithm}

\section{KubeLLM Implementation Details}

\subsection{Kubernetes Fault Resolution Dataset Construction}
\label{app:dataset_construction}

\textbf{Problem-Solution Pair Collection.}
We systematically gathered real-world Kubernetes issues and their corresponding solutions from authoritative technical forums, including Stack Overflow, GitHub issue discussions, and specialized Kubernetes blogs. Each collected sample represents an authentic troubleshooting scenario:

\begin{equation}
    \mathcal{D}_{\text{raw}} = \{(\rho_i, \varsigma_i) \mid \rho_i \in \mathcal{R}, \varsigma_i \in \mathcal{V}, i=1,2,...,N\}
\end{equation}
where $\mathcal{R}$ represents the set of Kubernetes problems, $\mathcal{V}$ denotes the corresponding solution space, and $N$ is the total number of raw samples collected.

\textbf{Problem-Attempt-Solution Reformulation.}
Through careful analysis of user interaction patterns in technical forums, we observed a distinctive characteristic of Kubernetes troubleshooting discourse: users typically document failed attempts before seeking external assistance. We formalized this observation by restructuring our data format to include attempted solutions:

\begin{equation}
    \mathcal{D}_{\text{struct}} = \{(\rho_i, \chi_i, \varsigma_i) \mid \rho_i \in \mathcal{R}, \chi_i \in \mathcal{X}, \varsigma_i \in \mathcal{V}\}
\end{equation}
where $\mathcal{X}$ represents the set of documented attempts made by users before arriving at the final solution. This three-component structure aligns with established internet troubleshooting etiquette, which encourages users to demonstrate prior effort before requesting assistance.

\textbf{Data Augmentation via Large Language Models.}
To expand our dataset beyond the limitations of manually collected samples, we employed the Grok 4 API to generate synthetic samples that maintain the established problem-attempt-solution structure:

\begin{equation}
    \mathcal{D}_{\text{aug}} = \mathcal{D}_{\text{struct}} \cup \{(\rho_j, \chi_j, \varsigma_j) \mid (\rho_j, \chi_j, \varsigma_j) \sim \pi_{\text{Grok}}(\cdot \mid \mathcal{D}_{\text{struct}})\}
\end{equation}
where $\pi_{\text{Grok}}$ represents the generation policy utilizing Grok 4 API, which was prompted to create new samples following patterns observed in our structured dataset.

\textbf{Semantic Deduplication.}
To ensure dataset quality and diversity, we applied sentence-transformer embeddings to perform semantic deduplication across our augmented dataset:

\begin{equation}
\begin{aligned}
    \mathcal{D}_{\text{dedup}} = \{&(\rho_k, \chi_k) \in \mathcal{D}_{\text{aug}} \mid \\
    &\forall j < k, \text{sim}(e(\rho_k, \chi_k), e(\rho_j, \chi_j)) < \tau_{dup}\}
\end{aligned}
\end{equation}

where $e(\cdot)$ denotes the embedding function provided by the sentence-transformer model, $\text{sim}(\cdot,\cdot)$ represents cosine similarity between embeddings, and $\tau_{dup}$ is the similarity threshold used to determine duplicate entries.

\textbf{Chain-of-Thought Enhancement.}
The final dataset transformation involved enhancing solution descriptions with explicit reasoning paths, transforming the intermediate problem-solution pairs into rich diagnostic examples:

\begin{equation}
    \mathcal{D}_{\text{final}} = \{(\rho_i, \chi_i, \varsigma_i, \varrho_i) \mid \varrho_i \sim \pi_{\text{GPT-5}}(\rho_i, \chi_i, \varsigma_i)\}
\end{equation}
where $\varrho_i$ represents the detailed reasoning process that connects the problem description and attempted solutions to the final resolution, generated using the GPT-5 API. This transformation was guided by three key objectives:

1. Enrichment of Kubernetes domain knowledge
2. Enhancement of diagnostic reasoning capabilities
3. Establishment of user-friendly output formatting conventions

\textbf{Dataset Composition.}
Through our comprehensive pipeline, we produced the final Kubernetes Fault Resolution Dataset comprising 7,000 high-quality data points. We strategically partitioned this dataset into two subsets: 5,000 samples for supervised fine-tuning (SFT) to establish foundational Kubernetes knowledge, and 2,000 samples reserved as a held-out test set for evaluating system effectiveness. This distribution ensures sufficient data for each training phase while maintaining a robust evaluation benchmark.

\textbf{Discussion on Synthetic Data Quality.}
While we acknowledge that synthetic data may exhibit quality disparities compared to purely real-world data, our approach specifically mitigates this concern through careful structural design. By maintaining the authentic problem-attempt-solution format observed in real troubleshooting scenarios, our synthetic data serves as valuable training material that exposes the model to diverse diagnostic pathways for similar problems. This diversity is particularly beneficial for training, as it enables the model to learn multiple valid approaches to resolving Kubernetes issues, mimicking the natural variation found in human problem-solving strategies. The structured format ensures that even synthetic examples maintain the logical flow and contextual richness characteristic of genuine troubleshooting interactions.

\subsection{KubeLLM Supervised Fine-Tuning Details}
\label{app:sft_details}

SFT adapts a pre-trained language model to the specialized domain of Kubernetes diagnostics through explicit supervision on labeled problem-solution pairs. The training dataset $\mathcal{D}_{\text{final}}$ consists of structured tuples $\{(\rho_i, \chi_i, \varsigma_i, \varrho_i)\}$, where each tuple contains a problem description, attempted solutions, final resolution, and explicit reasoning trace. This format enables the model to learn both domain knowledge and diagnostic reasoning patterns simultaneously.

To maintain computational efficiency while preserving the foundational capabilities of the base model, we implement Low-Rank Adaptation (LoRA) during fine-tuning. Rather than updating all parameters of the pre-trained model, we decompose weight updates into low-rank matrices:

\begin{equation}
W = W_0 + \Delta W = W_0 + A \cdot B^T
\end{equation}
where $W_0$ represents the frozen pre-trained weights, and the update matrices satisfy $A \in \mathbb{R}^{d \times r}$ and $B \in \mathbb{R}^{k \times r}$ with rank $r \ll \min(d,k)$. This decomposition significantly reduces the number of trainable parameters from $\mathcal{O}(d \cdot k)$ to $\mathcal{O}(r(d + k))$, enabling efficient adaptation while minimizing catastrophic forgetting.

The training objective optimizes the model's conditional probability distribution using maximum likelihood estimation:

\begin{equation}
\mathcal{L}_{\text{SFT}} = -\mathbb{E}_{(\rho,\chi,\varsigma,\varrho) \sim \mathcal{D}_{\text{final}}} \left[ \log p_\theta(\varrho,\varsigma \mid \rho,\chi) \right]
\end{equation}
where $\theta$ represents the trainable LoRA parameters. This formulation encourages the model to generate both accurate diagnostic reasoning paths and correct solutions when presented with Kubernetes problem descriptions and attempted remediation steps.

The SFT phase plays a central role in our methodology. Beyond injecting specialized Kubernetes diagnostic knowledge into the model, it significantly enhances structured reasoning capabilities and ensures robust alignment with real-world troubleshooting patterns present in the KFRD dataset. By establishing strong baseline competence in both technical accuracy and diagnostic reasoning structure, SFT effectively adapts the general-purpose Qwen3-8B model to the complexities of Kubernetes fault resolution while preserving its foundational language understanding. This careful adaptation process minimizes catastrophic forgetting of pre-trained knowledge, produces reliable and coherent diagnostic outputs from the outset, and creates a highly capable specialized model ready for direct deployment in the MetaKube architecture.

\section{KubeGraph Details}
\label{appendix:kubegraph}

\subsection{Overview}

KubeGraph is a comprehensive knowledge graph constructed using GraphRAG \cite{edge2024local} from diverse Kubernetes operational knowledge sources. The graph contains 44,022 entities and 111,832 relationships, structuring fault diagnosis knowledge into six primary categories. Through hierarchical clustering and community detection, KubeGraph organizes operational knowledge into interconnected subgraphs that enable efficient retrieval of relevant fault patterns and solutions, achieving 92\% precision in fault pattern matching with sub-second query response times.

\subsection{Knowledge Sources}

The knowledge corpus integrates three types of authoritative sources: (1) \textbf{Official documentation and community resources}, including Kubernetes official documentation and enterprise references from NewRelic and Alibaba Cloud, ensuring knowledge accuracy; (2) \textbf{Technical blogs and articles} from StackOverflow and Medium, incorporating real-world cases and solutions; (3) \textbf{Professional books} such as ``Kubernetes in Action'' and ``Kubernetes Practice Guide'', establishing systematic knowledge frameworks. The corpus underwent rigorous validation through cross-referencing and expert review to ensure reliability and practical applicability.

\subsection{Knowledge Categories}

KubeGraph organizes Kubernetes operational knowledge into six distinct categories based on fault domains and system components. Each category encapsulates specific error patterns, diagnostic procedures, and resolution strategies derived from production environments. This categorical organization, combined with GraphRAG's community detection capabilities, allows for efficient knowledge retrieval with an average query response time of 1.2 seconds across the entire knowledge base.

\begin{table}[h]
\centering
\small
\begin{tabular}{p{0.22\linewidth}p{0.68\linewidth}}
\toprule
\textbf{Category} & \textbf{Focus Areas} \\
\midrule
Resource Errors & OOMKilled events, CPU throttling, resource quotas, PVC mounting failures, memory leaks, autoscaler misconfigurations \\
Network Errors & Service discovery failures, DNS resolution, network policies, ingress configuration, load balancer issues, CNI problems \\
Scheduling Errors & Node affinity violations, taint/toleration mismatches, insufficient resources, pod priority conflicts, DaemonSet placement \\
Image Errors & ImagePullBackOff, registry authentication, rate limiting, private registry access, multi-arch compatibility, layer corruption \\
Configuration Errors & ConfigMap/Secret mounting, RBAC permissions, admission webhooks, environment variables, Helm values, operator configs \\
System Errors & Container runtime failures, kubelet crashes, etcd issues, certificate expiration, kernel panics, filesystem corruption \\
\bottomrule
\end{tabular}
\caption{Knowledge categories and focus areas in KubeGraph}
\end{table}

\end{document}